# Effects of term weighting approach with and without stop words removing on Arabic text classification

1st Esra'a Alhenawi
*Department of Software Engineering*
*Faculty of Information Technology*
*Al-Ahliyya Amman University*
Amman, Jordan
e.alhenawi@ammanu.edu.jo

2nd Ruba Abu Khurma
*Department of Computer Science*
*Faculty of Information Technology*
*Al-Ahliyya Amman University*
Amman, Jordan
r.khurma@ammanu.edu.jo

3rd Pedro A. Castillo, Maribel G. Arenas
*University of Granada*
Granada, Spain
pacv@ugr.es,
mgarenas@ugr.es,

*Abstract*—Classifying text is a method for categorizing documents into pre-established groups. Text documents must be prepared and represented in a way that is appropriate for the algorithms used for data mining prior to classification. As a result, a number of term weighting strategies have been created in the literature to enhance text categorization algorithms' functionality. This study compares the effects of Binary and Term frequency weighting feature methodologies on the text's classification method when stop words are eliminated once and when they are not. In recognition of assessing the effects of prior weighting of features approaches on classification results in terms of accuracy, recall, precision, and F-measure values, we used an Arabic data set made up of 322 documents divided into six main topics (agriculture, economy, health, politics, science, and sport), each of which contains 50 documents, with the exception of the health category, which contains 61 documents. The results demonstrate that for all metrics, the term frequency feature weighting approach with stop word removal outperforms the binary approach, while for accuracy, recall, and F-Measure, the binary approach outperforms the TF approach without stop word removal. However, for precision, the two approaches produce results that are very similar. Additionally, it is clear from the data that, using the same phrase weighting approach, stop word removing increases classification accuracy.

## I. INTRODUCTION

The technique of classifying a given text into one or more categories is known as text classification (TC). Due to the fact that a training set of labeled (pre-classified) documents is provided, this procedure is regarded as a supervised classification technique. The purpose of TC is to use a certain classifier to give a label or choose the correct category for an unlabeled text item.

Due to the rapid expansion of the web and computer innovations, there are now billions of electronic text documents that are developed, modified and archived digitally. As a result, finding, organizing, and storing these documents has become extremely difficult for the general public, particularly computer users. For this reason, categorization of text is crucial and can be used in a variety of applications, including the categorization of stories from the media, messages sent via email (spam filtering) where email messages are classified into the two categories spam and non-spam.

There have been numerous studies on categorizing English documents using various classification techniques, including: Bayes's naive classifier (Al-Salemi and Aziz, 2011), K-Nearest Neighbor (KNN) (He et al., 2003), neural networks (Peng et al., 2003), and Support Vector Machine (AVM) (Ruiz and Srinivasan, 1999).

Despite the fact that there hasn't been much research on Arabic text classification contrasted to the studies that have been conducted regarding text classification for languages other than Arabic, a lot of investigation has gone accomplished in this area over the past decade. The text classification strategies used include decision trees, SVM, rule induction, neural networks, and statistical techniques.

Arabic is an extremely rich language with complicated morphology, making Arabic Text Classification (ATC) a challenging undertaking. Arabic is a Semitic language with 28 letters and a total of 25 consonants and three long vowels. Diacritics are signals positioned either below or above characters to amplify the letter in pronunciation, such as dama, fatha, Shada, kasra, and double kasra. Arabic is written from right to left, and Arabic letters have different styles depending on the letter position (beginning, middle, or end of word). However, the majority of studies on Arabic text classification employ documents without diacritics or papers without diacritics since dealing with diacritics requires taking into account Arabic grammar and syntax, which is a highly challenging issue. The Arabic language allows for the possibility of several meanings being attached to the same lexeme depending on the situatio as shown in Table I.

| Arabic Word | Arabic Mean | English Mean |
|---|---|---|
| H @ Al ' „ | an4ı H @ Al | Correctness itself |
| tn'B „ | ,tA'B@ A | Eye |

TABLE I: Example of HOMONYMY in Arabic Language.

The three word classes in Arabic are nouns, verbs, and particles, albeit a word may fall under more than one lexical category (noun, verb, adjective, etc.). in various circumstances, as in the example in Table III.

|  | Arabic Mean | English Mean |
|---|---|---|
| ذهب / ar,z,t@ | ذهب | Go to (verb) |
| ذهب | ذهب | Gold (Noun) |

TABLE II: Word Tags AMBIGUITY in Arabic Language.

Also, the basic sentence structure for Arabic is very flexible and may take the form of "Subject Verb Object"

الولد أكل التفاحة

, "Verb Subject Object"

أكل الولد التفاحة

or "Verb Object Subject"

أكل التفاحة الولد

so we can't find the name based on sentence analysis as in the English language. Arabic has many words that are considered synonymous such as

أعطى، منح، وهب، 

which means ( give ) in English, The form of some Arabic words may change according to their case modes (nominative, accusative, or genitive) such as the plural of the word

متفائل

which means Optimistic be the form

متفائلون

in the case of nominative

متفائلين

and the form the word

متفائلين

in the case of accusative/genitive

المتفائلون

where Arabic light stemming can handle these cases.

The size of the feature vectors retrieved from the text grows as a result of the richness of the Arabic language, but thankfully, Arabic has a built-in filtering mechanism that allows words to be mapped into their root patterns using the stemming tools that were covered in the stemming process.

The processes of Arabic text classification begin with the gathering of Arabic text documents (corpus collection), followed by the preprocessing step, which entails three stages: stop words removal, stemming, and feature extractions. The next stage is to split these texts into two sections, one for training (to learn the classifier) and the other for testing (to determine the classifier's accuracy).

According to machine learning (ML), a set of documents (training set) will be used to build the classifier by processing the training documents for each category again and by a variety of Information Retrieval (IR) techniques to extract a set of features used as characteristics for each category, while the test set (used to test and evaluate the classifier's accuracy) will be used to classify the documents under each category as "unseen documentation." Classification of Arabic text passes through the four stages depicted in figure 1.

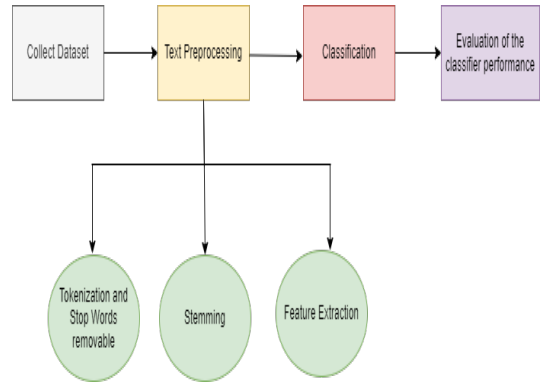

Fig. 1: Arabic Text Classification Stages

There are numerous approaches to classification, but in this study, we used statistical methods such as the Multivariate Bernoulli Naive Bayes Classifier (MBNB), a variation of the Naive Bayes classifier that has been used in tasks involving text classification that take the presence or absence of document words into account, and the Multi-nominal Model, which takes the frequency of document words into account.

This study compares the impact of term weighting approaches (binary vs. term frequency (TF)) on Arabic classification outcomes based on accuracy, recall, precision, and f-measure metrics with and without stop word removal. To do this, we ran four tests, two of which included stop word removal, first using a binary term weighting technique, then a TF term weighting approach, and the other two of which did not.

The rest of the paper is organized as follows: sec. II reviews some of the studies regarding text categorization for the Arabic language. Sec.III presents the proposed methodology.

## II. LITERATURE REVIEW

Recent research have used a variety of methods and metrics to assess the effectiveness of various methods in addressing the challenge of text classification. The text categorization field for the Arabic language is covered in this section through a number of studies and tests.

The impact of several word weighting strategies, such as TF, TF-IDF, and TF-ICF, on the classification of Arabic text was compared by Guru et al. in 2019. With a KNN classifier, they evaluated a dataset with more than 5300 Arabic complaints. The findings showed that there is a stability difference between the word weighting approaches that were investigated. (Guru et al., 2019).

Arabic text materials were automatically categorised using NB by El-Kourdi et al. in their publication (El Kourdi et al., 2004). The stated accuracy ranged from 68.78% on average and 92.8% at its highest. El-Kourdi employed a corpus of 1500 text documents divided into 5 categories, with 300 texts in each group. The roots of each word in the papers are converted. The resulting corpus has a vocabulary of 2,000 terms/roots. The evaluation employed cross-validation.

Rehab M. Duwairi (Duwairi et al., 2007) examined the three classification methods of distance-based classifier, KNN classifier, and Naive Bayes classifier on a collection of 1000 texts that were divided into 10 groups (100 documents per category) and varied in length and writing style. These documents were split in half, with one half being used for training and the other for testing. The words' roots were taken out using a stemmer that operates on roots. To gauge the classifiers' accuracy, recall, precision, error rate, and fallout were taken into consideration. The studies' findings demonstrated that the Naive Bayes classifier performed better than the other two. Mohamed El Kourdi et al. employed the root-based stemmer to extract the word roots before classifying Arabic web articles using the Naive Bayes classifier.

Utilizing the Bayesian theorem as its foundation, Bassam Al-Salemi and Mohd. Juzaiddin Ab-Aziz (Al-Salemi and Aziz, 2011) compared the performance of three classifiers: the Simple Nave Bayes (NB), Multi-variant Bernoulli Nave Bayes (MBNB), and Multinomial Nave Bayes (MNB) models. Arabic stemming was done using the TREC-2002 Light Stemmer. As a result, the MBNB classifier performs better than both the NB and MNB classifiers.

A method for Arabic multi-label classification based on the Binary Relevance approach was proposed by Taha and Tiun (Taha and Tiun, 2016) in their paper Taha2016binary. SVM, KNN, and Naive Bayes (NB) are just a few examples of the various machine learning classifiers that have been employed. Additionally, three feature selection techniques (Odd ratio, Chi-square, and Mutual information) were added to improve the multi-label classification process' accuracy. 10,000 items in 5 categories of Modern Standard Arabic (MSA) are evaluated. With an overall F-measure of 86.8% for the multi-label classification of Arabic text, the results demonstrate a considerable influence of the feature selection methods utilized on the classification.

To deal with large amounts of data the idea of Parallel processing was deployed which implies multiple processors in a cluster to ensure reasonable delay when building the classifier. Mona and Shurug (Alshahrani and Alkhalifa, 2018) developed a tool called Parallel Arabic Text Classifier (PATC) for Arabic text classification using a parallel programming framework. PATC is composed of three phases; Preprocessing phase to normalize and stem the Arabic corpus. The second phase is training or Building the Classifier, and lastly is the testing or Classifying phase. To build the classifier MapReduce distributed programming model is used to associate each label with frequent words. The accuracy of the classification was around 80% using single-label measures and 90% using multi-label measures.

The concept of parallel computation, which assumes numerous processors in a cluster to ensure a fair latency when generating the classifier, was introduced to deal with massive volumes of data. Using a parallel programming framework, Mona and Shurug (Alshahrani and Alkhalifa, 2018) created the Parallel Arabic Text Classifier (PATC), a tool for classifying Arabic text. Three stages make up PATC. The first stage normalizes and stems the Arabic corpus. The testing or classifying step comes after the second phase of training or building the classifier. Each label is linked to frequent terms using the MapReduce distributed programming methodology to construct the classifier. When utilizing single-label measures, the classification accuracy was about 80%, and when using multi-label measures, it was 90%.

Saad (Saad, 2010) provided an Arabic corpus called "OSAC" that was compiled from various websites and contains 22.429 documents divided into ten categories (history, economics, entertainment, religious and fatwas, education & family, sports, astronomy, stories, low, health, cooking recipes), and made it freely accessible online for researchers to use. Following the removal of stop words, he compared the accuracy of roughly eight classifiers, including the C4.5 Decision Tree (TD), KNN, and SVMs with light stemming and root stemming techniques. Later, he examined the accuracy of these classifiers under other term weighting approaches, although in all of his studies, stop words were eliminated.

By using binary term weighting first, followed by TF term weighting approach, with stop word removing in two experiments, and without in the remaining two experiments, we hope to investigate the effects of two preprocessing steps (term weighting approach and stop word removal) in the Arabic text classification process for NB variant classifiers. Accuracy, recall, precision, and F_Measure were used to gauge the classifier's effectiveness.

III. METHODOLOGY

*A. Collect an Arabic documents (corpus)*

Corpus in this paper consists of 322 documents which are categorized into six main topics: Agriculture, Economy, Health, Politic, Science, and Sport. Each category is 50 documents except the Health category which contains 61 documents (Arabic, )

*B. Text preprocessing*

In this step, documents are processed and prepared to be used in the classification phase. This phase has many sub-phases: Tokenization, text cleaning(stop words removing), Stemming, and feature extracting based on terms weighting.

1) Tokenization: this sub-step is used for tokenizing the document word by word.
2) Text cleaning (stop words removing): removes the non-Arabic letters, numbers, and adverbs

days of week

, the month of the year

punctuation's ( ,mP'," l HtAC )

pronouns ( ,It Al@ )

, conjunctions ( abel e9 >)

and prepositions ( ,>I @ e9 > )

and others. This step was carried out in two of the four experiments that we conducted to compare the effect of the term weighting approach (binary or term frequency) on the classification with and without passing this sub-phase in preprocessing phase.

3) Stemming: this crucial preprocessing step uses a method to lower the high dimensionality of the feature vectors used in text classification. Working with Arabic document words without stemming results in a huge number of words being input into the classification phase, which will increase the classifier complexity and have an impact on the efficiency and accuracy rate for this classifier. The Arabic language is highly derivative, where many hundreds of words could be formed using only one stem, in addition to the fact that a single word may be derived from multiple stems. Since words can be mapped into their root patterns using the Arabic language's built-in filtering mechanism, the stemming step is crucial. (stemmer, ).

4) The most informative terms are extracted from documents depending on their weighting in the feature extraction process. The primary goal of term weighting is to improve the feature vector representation of text documents. The Boolean model, which signals the absence or existence of a word with Boolean 0 or 1 correspondingly, is one of the widely used term weighting techniques. The count model, which displays the frequency of each word in the text, is another.

The first advantage of feature extraction is that it minimizes the number of dimensions (terms), which in turn reduces the complexity of the classifier and the processing needs (time, memory, and desk space). The second advantage is an improvement in classification efficiency due to the removal of noise features and the prevention of overfitting brought on by phrases that are common across all categories.

This paper aimed to compare the effects of the previous two-term weighting approaches on the classification process with and without stop word removal.

5) Classification: results of the previous steps used as input for this step. In this paper, we used two Naïve Bayes flavors classifiers.

6) Evaluation: for evaluating the performance of classifiers in each experiment we used accuracy, recall, precision, and F-measure matrices. [13]

    - Accuracy: it is the percentage of correct predictions that had been made by the classifier.

    - Recall: is a statistic that expresses the probability that a classier can accurately anticipate positive occurrences from all actual positive instances. Even if it is successfully anticipated as positive data points or incorrectly projected as negative data points, it is equal to the ratio between correctly classified as positive (TP) data points and a sum of positive data points. (TP + FN).

    - Precision: is a statistical measure of the proportion of positive instances that a classifier can accurately predict out of all the positive examples that it predicts. Regardless of whether a data point is actually accurately positive or not, it is equal to the ratio between data points that are correctly identified as positive (TP) and the total of data points that are projected to be positive. (TP + FP).

    - F-measure: it is a geometric mean of recall and precision.

IV. RESULTS

Four experiments were conducted, with the first experiment using the TF term weighting approach with stop words removed, the second using the binary term weighting approach with stop words removed, the third using the binary term weighting approach with stop words removed, and the fourth using the TF term weighting approach without stop words removed.

Table III illustrated a comparison results between the TF term weighting approach vs. the binary approach in term weighting based on accuracy, recall, precision, and F-Measure with stop words removed at first and without next.

TABLE III: FOUR EXPERIMENTS RESULTS.

| Experiment | Accuracy | Recall | Precision | F-measure |
|---|---|---|---|---|
| Exp.1 | 0.953 | 0.953 | 0.960 | 0.954 |
| Exp.2 | 0.922 | 0.922 | 0.940 | 0.923 |
| Exp.3 | 0.882 | 0.882 | 0.914 | 0.877 |
| Exp.4 | 0.913 | 0.937 | 0.913 | 0.913 |

From table III, we discovered that the TF weighting approach provides better accuracy, recall, precision, and F-Measure than the binary approach with stop words removed. On other hand without stop words removing the binary approach in term weighting provides better values than TF.

Figure 2 presents accuracy results where the TF term weighting approach outperformed the binary term weighting approach with stop words removing by 3% while without stop words removing binary approach provides better accuracy than TF approach with a percentage equal 4%.

Furthermore, the binary term weighting strategy offers a higher Recall value than the TF term weighting approach, with a score equal to .953 without stop words deleted, while the binary approach offers a higher Recall value than T, with a score equal to 0.937 without stop words removed TF.

With a score of 0.960, the binary term weighting technique with stop words removed performs better in terms of precision than the TF term weighting approach, while the scores for the two approaches without stop words removed are practically equal. 3.

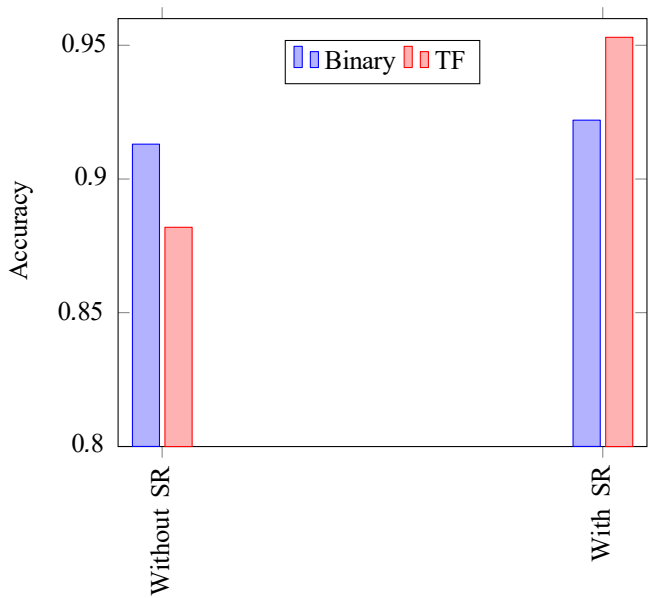

Fig. 2: Four experiments results based on Accuracy.

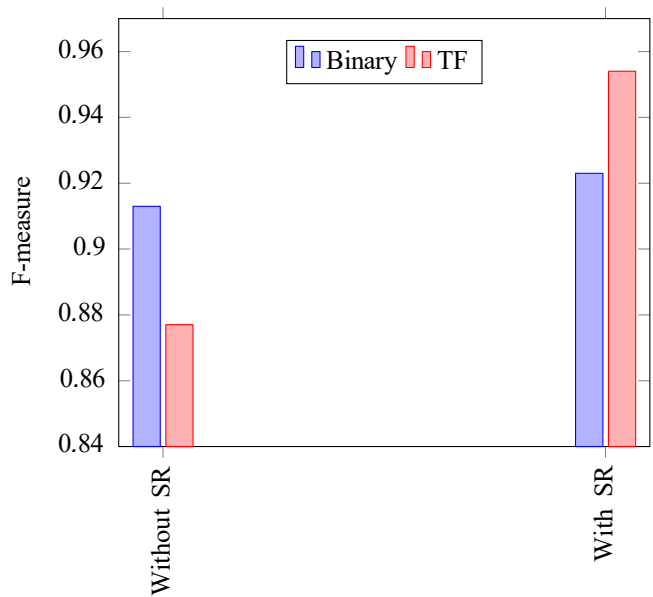

Fig. 4: Four experiments results based on F-measure.

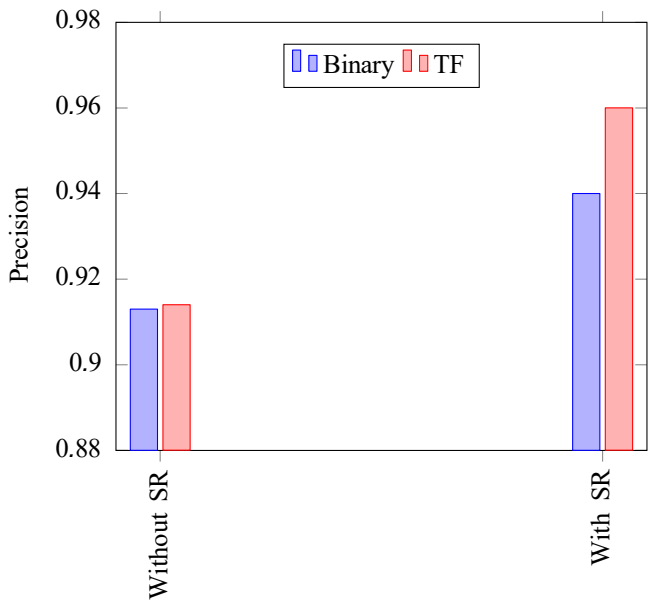

Fig. 3: Four experiments results based on Precision.

F-Measure for Binary term weighting approach outperformed the TF term weighting approach with a score equal to 0.913 without stop words removed however with stop words removed TF approach outperformed the binary approach with a score equal to .954 which is expected as F_Measure metric value depends on recall and precision metrics values, as shown in figure 4.

## V. Conclusion

In this study, we examined the effects of two feature weighting approaches on the text classification (TC) process with and without stop word removal: the Binary and Term frequency approaches. We used a corpus of 322 Arabic papers divided into six primary categories (Agriculture, Economy, Health, Politics, Science, and Sport), each of which has 50 documents (with the exception of the Health category, which contains 61 documents).

The findings demonstrated that, for the identical term weighting technique, stop word removal improved the classifier's accuracy, recall, and precision. Without stop words removed, the binary weighting technique outperformed the TF approach, whereas with stop words removed, the TF approach outperformed the Binary approach.

## Acknowledgements

This work was supported by the project number PID2020-115570GB-C22 (DemocratAI::UGR) and by the Catedra de Empresa Tecnologia para las Personas (UGR-Fujitsu).